\documentclass{ai4x}
\usepackage{tabulary}

\title{Beyond Semantic Similarity: A Two-Phase Non-Parametric Retrieval Workflow for Corporate Credit Underwriting}

\hypersetup{
    pdfkeywords={AI, machine learning, science, conference template}
}

\begin{document}
\twocolumn[
\maketitle
\icmlsetsymbol{equal}{*}
\begin{icmlauthorlist}
\icmlauthor{Linus Ng Junjia}{ocbc}{}{presenting}
\icmlauthor{Ezekiel Tee Kongquan}{ocbc,gatech}{}{}
\icmlauthor{Kelvin Heng}{ocbc}{}{}
\icmlauthor{Kenneth Zhu Ke}{ocbc}{}{}
\icmlauthor{Zhao Jing Yuan}{ocbc}{}{}
\end{icmlauthorlist}

\icmlaffiliation{ocbc}{OCBC, Singapore}
\icmlaffiliation{gatech}{Georgia Institute of Technology}
\icmlcorrespondingauthor{}{linus.ng@ocbc.com}
\icmlcorrespondingauthor{}{ezekieltee@ocbc.com}
\icmlcorrespondingauthor{}{kelvinheng@ocbc.com}
\icmlcorrespondingauthor{}{kennethzhu@ocbc.com}
\icmlcorrespondingauthor{}{jingyuanzhao@ocbc.com}

\printAffiliations
\vskip 2.5ex
]
\thispagestyle{fancy}

\section{Introduction}
\label{sec:introduction}
Corporate credit underwriting relies heavily on the analysis of long-form financial documents such as annual reports and industry reports \cite{hybridrag-financial-docs-2024}. Analysts must extract relevant financial indicators, assess risk disclosures, and synthesize insights from documents that can span hundreds of pages and multiple languages \cite{hybridrag-financial-docs-2024}.

Retrieval-Augmented Generation (RAG) systems have emerged as a promising approach to assist document-intensive workflows \cite{lewis2020rag,gao2024retrievalaugmentedgenerationlargelanguage}. By retrieving supporting passages from external corpora, RAG systems can improve factual grounding and reduce hallucination in language model outputs \cite{lewis2020rag,gao2024retrievalaugmentedgenerationlargelanguage}. However, standard RAG pipelines typically prioritize semantic similarity between queries and document passages \cite{lewis2020rag}. In financial analysis tasks, this objective often fails to align with the needs of analysts \cite{hybridrag-financial-docs-2024}.

Financial documents often contain narrative descriptions, regulatory disclosures, and repetitive boilerplate language \cite{hybridrag-financial-docs-2024}. As a result, similarity-based retrieval systems may surface passages that share terminology with the query but lack actionable analytical value \cite{utility-focused-llm-2025}. We refer to this issue as the \textit{similarity–utility gap} \cite{utility-focused-llm-2025}.

To address this challenge, we propose a retrieval architecture designed specifically for enterprise decision-support workflows \cite{utility-focused-llm-2025}. 
Our system introduces a two-phase pipeline that separates high-recall retrieval from high-precision utility ranking \cite{utility-focused-llm-2025}. 
The architecture incorporates hybrid lexical-semantic retrieval, adaptive candidate refinement, and a utility-grounded ranking framework in which a language model evaluates passages according to their analytical usefulness \cite{hybridrag-financial-docs-2024, utility-focused-llm-2025}.

The system is designed for deployment in regulated financial environments and operates entirely on-premise using self-hosted open-source models. This ensures compliance with strict data governance policies while maintaining high retrieval performance.

Our main contributions are:

\begin{itemize}
\item A utility-grounded retrieval framework that aligns passage ranking with decision usefulness in credit underwriting tasks.
\item An adaptive retrieval controller that filters candidate passages using query intent and document structure signals.
\item A context-aware extraction module that preserves structural information in narrative and tabular financial content.
\item An enterprise deployment demonstrating large productivity gains for analysts.
\end{itemize}

\section{Related Work}

Retrieval-Augmented Generation has become a widely adopted approach for grounding large language models in external knowledge sources \cite{lewis2020rag,gao2024retrievalaugmentedgenerationlargelanguage}. Early work introduced dense retrieval methods using neural embeddings to retrieve semantically relevant passages from large corpora \cite{lewis2020rag}. Hybrid retrieval approaches combining lexical search with dense embeddings have further improved recall in heterogeneous document collections \cite{santhanam2022colbertv2effectiveefficientretrieval}.

Recent research has explored the use of language models as evaluators for ranking retrieved content \cite{zheng2023judgingllmasajudgemtbenchchatbot}. These \textit{LLM-as-a-Judge} approaches leverage the reasoning capabilities of language models to assess the quality and relevance of candidate passages \cite{zheng2023judgingllmasajudgemtbenchchatbot}. Such methods have been applied in question answering, information retrieval evaluation, and ranking tasks.

In enterprise settings, RAG systems must also address constraints related to data governance, privacy, and auditability. Financial and legal institutions often require on-premise deployment and traceable source attribution, which introduces additional design considerations beyond model performance.

Our work contributes to this literature by proposing a retrieval architecture that explicitly optimizes for analytical utility in enterprise workflows. By combining hybrid retrieval, adaptive candidate control, and utility-based ranking, the system prioritizes passages that contain actionable financial evidence rather than merely semantically similar text.

\setlength{\parskip}{0pt} 
\section{Methodology}

\subsection{Problem Setting}

Corporate credit underwriting requires analysts to justify reported financial statements derived from long, heterogeneous financial documents such as annual reports and industry reports. These documents often contain dense narrative sections, multilingual commentary, and unstructured financial tables. Traditional RAG pipelines retrieve passages based primarily on semantic similarity, which frequently results in content that is topically related but not useful for decision-making.

We formalize the task as retrieving and ranking a set of document segments that maximize decision utility rather than semantic similarity. Given a user query \(q\) and a corpus of long financial documents \(D=\{d_1, d_2, ..., d_n\}\), the objective is to identify a set of passages \(P^* \subset D\) that contain verifiable evidence relevant to credit underwriting decisions.

Our system introduces a two-phase retrieval and re-ranking architecture designed to bridge the gap between semantic similarity and decision utility while operating entirely within an on-premise environment.

\subsection{System Overview}

The proposed architecture consists of five main components: document ingestion, hybrid candidate retrieval, adaptive retrieval controller, utility-grounded re-ranking, and context-aware evidence extraction.

The overall pipeline is illustrated conceptually as:
\[
q_{\text{statement}} \rightarrow R_{\text{hybrid}}(D) \rightarrow C_{\text{adaptive}} \rightarrow J_{\text{utility}} \rightarrow E_{\text{context}}
\]
where \(q_{\text{statement}}\) a query supplemented with its corresponding financial statement, \(R_{\text{hybrid}}\) retrieves an initial candidate pool, \(C_{\text{adaptive}}\) filters candidates using query-aware reasoning, \(J_{\text{utility}}\) ranks candidates by decision usefulness, \(E_{\text{context}}\) extracts the final evidence


\subsection{Document Ingestion}

Corporate financial documents are segmented into structured sections using document layout cues. Each segment is indexed with metadata that include the source of the document, the title of the section, and the page references. This preprocessing step ensures that downstream components can leverage structural information during retrieval and extraction.

\subsection{Phase 1: Hybrid Candidate Retrieval}

The first phase performs broad retrieval to maximize recall across multilingual and heterogeneous financial documents.

Given a query \(q\), we retrieve an initial candidate set \(C_0\) using a hybrid retrieval strategy that combines keyword retrieval and dense semantic retrieval using multilingual embeddings:



\[
C_0 = \text{TopK}_{\text{kw}}(q, D) \cup \text{TopK}_{\text{embed}}(q, D)
\]



In Hybrid retrieval, keyword retrieval preserves precision for financial terminology, while semantic retrieval captures paraphrased or contextual references.

The choice of the TopK value represents a critical hyperparameter at this stage: setting it too low risks restricting the candidate pool passed to the subsequent phases, while setting it too high diminishes the value of initial retrieval. 
With a TopK value set to 50, the result is a high-recall candidate pool of passages for further processing.

\subsection{Phase 2: Adaptive Retrieval Controller \& Utility-Grounded Re-ranking}

Not all retrieved passages are useful for credit analysis. Financial reports contain boilerplate disclosures, legal notes, and narrative sections that may be semantically related to a query, but irrelevant to underwriting decisions.

To address this issue, we introduce an adaptive retrieval controller that evaluates candidate passages using query intent and document structure.

Given candidate passages \(C_0 = \{p_1, ..., p_k\}\), the controller predicts relevance and support for a given passage:
\[
Rel_i = f(q_{\text{statement}}, p_i, m_i)
\]
\[
S_i = f(q_{\text{statement}}, p_i, m_i)
\]
\[
U_i = f(q_{\text{statement}}, p_i, m_i)
\]

where:

\begin{itemize}
    \item \(q_{\text{statement}}\) is a query supplemented its corresponding financial statement
    \item \(p_i\) is the candidate passage
    \item \(m_i\) represents structural metadata
    \item \(Rel_i\) represents relevancy (boolean) of candidate passage
    \item \(S_i\) represents evidence support (boolean) of candidate passage
    \item \(U_i\) represents utility score (numerical score) of candidate passage
\end{itemize}

The controller is implemented using a lightweight language model that evaluates whether a passage is likely to contain relevant and supportive information.

This stage produces a refined candidate set:
\[
C_1 = \{p_i \in C_0 \mid (S_i) \cdot [Rel_i] \}
\]

where \(S_i \cdot [Rel_i]\) represents the logical condition that \(S_i\) is effectively nullified unless \(Rel_i\) succeeds.

This mechanism is conceptually related to adaptive retrieval strategies that condition retrieval on model reasoning rather than fixed pipelines \cite{asai2023selfraglearningretrievegenerate}.


The remaining candidates are ranked according to decision utility using an LLM-as-a-Judge framework.
Rather than measuring similarity to the query, the judge evaluates each passage based on its usefulness for underwriting decisions. Given a passage \(p_i\), the judge produces a utility score:



\[
J_1 = \{p_i \in C_1 \mid U_i \geq U_{\text{threshold}}\}
\]





where \(U_i\) is the utility score and \(U_{\text{threshold}}\) is the tunable utility score threshold.
\(U_{\text{threshold}}\) serves as a critical hyperparameter in the architecture, balancing precision and recall in the final evidence set. 
A lower \(U_{\text{threshold}}\) increases recall by admitting a larger set of passages, which is advantageous for exploratory queries or scenarios where analysts require broad and comprehensive coverage. 
Conversely, a higher \(U_{\text{threshold}}\) prioritizes precision, returning only passages with the highest decision utility, which is ideal for targeted queries where concise evidence is needed.

The utility-grounded re-ranking mechanism enables the system to prioritize passages containing financial indicators and industry signals.

\subsection{Context-Aware Evidence Extraction}

Financial documents frequently contain complex tables, footnotes, and structured subsections. Simple chunk extraction can distort meaning, break structural relationships, or cause loss of attribution.
We therefore introduce a context-aware extraction module that dynamically selects the appropriate extraction strategy based on document structure.

Two extraction modes are employed depending on the structure of the source content. For narrative sections, relevant text spans are extracted using markdown-aware segmentation, which preserves structural elements such as section headers, bullet lists, and paragraph boundaries. 

When information appears in tables or structured financial statements, the system distinguishes between complex and non-complex tables. Non-complex tables (tables with single-level headers and regular grid structures) are parsed to extract the relevant rows or cells. In contrast, complex tables containing multi-level headers, hierarchical indices, merged cells, or irregular layouts are preserved along with source metadata, including the document name and page reference, to support manual verification and accurate source attribution.

Two extraction modes are used:

\begin{itemize}
    \item \textbf{Localized Passage Extraction}
    
    For narrative sections, relevant text spans are extracted using markdown-aware segmentation.
    
    This preserves structural markers such as:
    
    \begin{itemize}
        \item section headers
        \item bullet lists
        \item paragraph boundaries
    \end{itemize}
    
    \item \textbf{High-Fidelity Table Citation}
    
    When information resides within tables or structured financial statements, the system classifies tables as either complex tables or non-complex tables:
    
    \begin{itemize}
        \item \textbf{Complex tables}: Multi-level headers, hierarchical row indices, irregular structures, or merged cells that require specialized handling
        \item \textbf{Non-complex tables}: Single-level headers and row indices, well-structured grids with regular formatting
    \end{itemize}
\end{itemize}

For non-complex tables, the system performs structured table parsing to extract the relevant rows or cells. For complex tables, rather than attempting to parse the structure, the system preserves the full table context, including source metadata such as document name, and page reference. This information is appended as supplementary information to support manual verification and referencing.

This approach ensures that extracted financial metrics remain interpretable and can be reliably traced back to their original document context, while accommodating the diversity and structural complexity of tables across different reports.










\section{System Architecture}

The proposed system consists of five main components: document ingestion, hybrid retrieval, adaptive candidate control, utility-grounded re-ranking, and context-aware extraction.

\begin{figure*}[t]
\centering
\includegraphics[width=\linewidth]{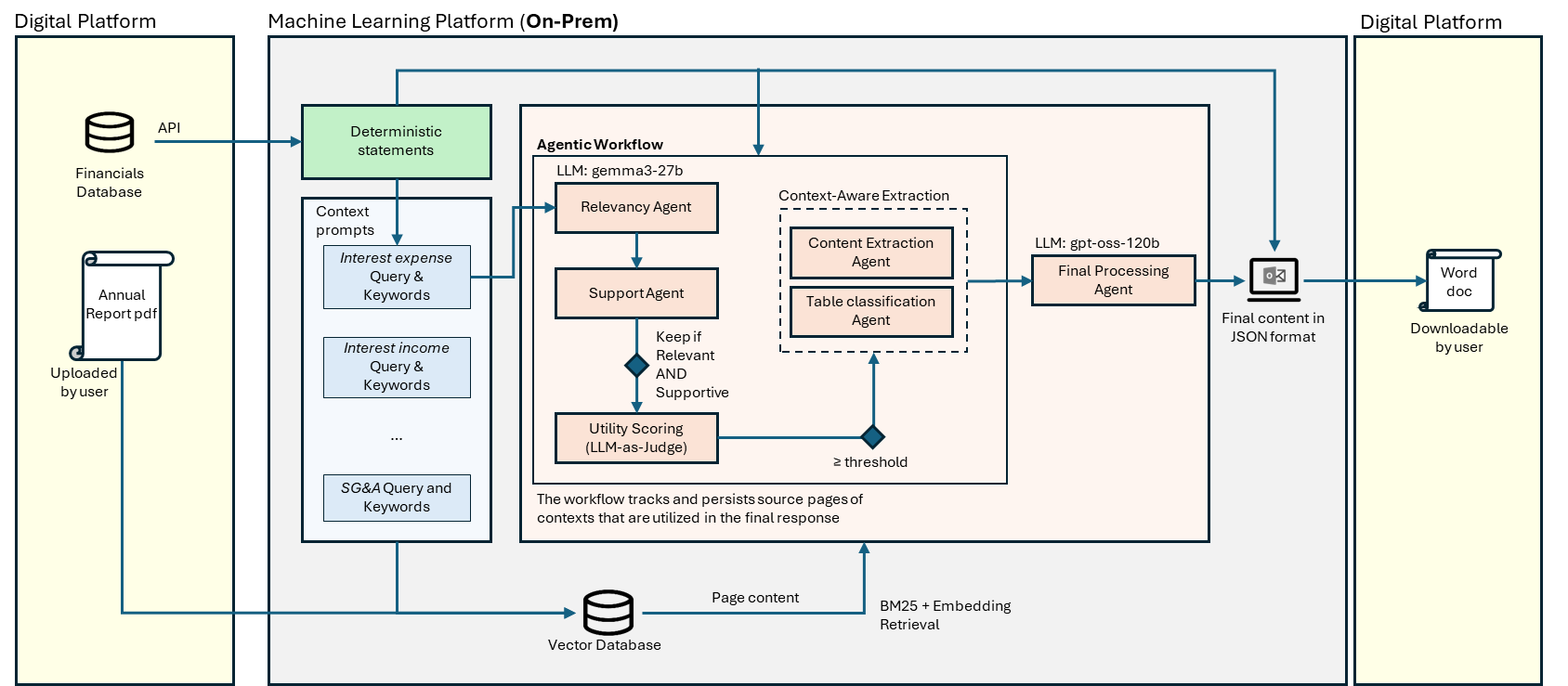}
\caption{Utility-grounded retrieval architecture for long-document financial analysis.}
\label{fig:architecture}
\end{figure*}

Figure~\ref{fig:architecture} illustrates the overall pipeline.

To satisfy regulatory and data governance constraints in corporate credit underwriting, the proposed system is deployed entirely within an on-premise environment, with all components executed under enterprise-controlled infrastructure. This ensures that sensitive financial data remains local across all stages of the pipeline.

The deployment architecture is structurally aligned with the proposed formulation defined in Section 3.2 and is realized through a collection of self-hosted modules that implement each stage without reliance on external services.

\begin{itemize}
    \item Retrieval Infrastructure
    
    Hybrid retrieval is implemented via the combination of lexical and dense retrieval mechanisms over locally indexed document collections. Keyword-based retrieval operates on structured indices, while dense retrieval is enabled through multilingual embedding models deployed on-premise. Document representations are stored and queried through an internal vector database, facilitating efficient construction and propagation of the candidate set \( C_0 \).
    
    \item Agentic Reasoning Modules
    
    The adaptive retrieval controller and utility-grounded re-ranking stages are instantiated using self-hosted language models. Lightweight language models are utilized for relevance and support classification to enable efficient pruning of candidates, while higher-capacity models perform utility scoring under the LLM-as-a-Judge framework and support final response generation. This separation reflects the staged evaluation functions defined in Section 3.5, while maintaining computational tractability within an on-premise setting.
        
    \item Context-Aware Extraction
    
    The evidence extraction stage operates directly on the filtered and ranked candidate set \( J_1 \), applying structure-aware processing to heterogeneous document segments. Narrative text and tabular content are handled through distinct extraction strategies executed locally, preserving structural fidelity and source attribution without external transformation.

    \item Data Flow and Persistence
    
    System components are coordinated through internal messaging pipelines that support scalable ingestion and processing of financial documents. Intermediate artifacts (candidate passages, filtering decisions, utility scores, and associated metadata) are persistently maintained, enabling reproducibility, traceability, and auditability of all outputs.
        
\end{itemize}


This design ensures that all decision-relevant computations remain verifiable, controllable, compliant with enterprise governance requirements, and that sensitive financial data never leaves the enterprise infrastructure.





\section{Results}

We evaluated the system on a multilingual corpus of financial documents with relevance labels curated by credit analysts.
Compared to traditional naive retrieval systems, the proposed approach significantly improves retrieval performance.

Due to the proprietary nature of the financial documents used in this study, the evaluation was conducted in a restricted enterprise environment. The dataset consists of internal corporate credit documents including financial reports, industry analyses, and related underwriting materials used in production workflows. These documents contain confidential financial information and cannot be publicly released. As a result, the annotated dataset and detailed case studies used for evaluation are not included in the public version of this work.

To ensure meaningful evaluation despite these constraints, financial statements and additional contexts were curated by senior credit analysts based on real underwriting tasks. Queries were constructed to reflect analytical questions commonly encountered in credit assessment workflows, such as identifying components in balance sheets, income statements, and industry outlook signals.

In real-world deployment across more than 800 analysts, the system reduced document review time from several hours to approximately three minutes.

\section{Summary}

The results highlight the limitations of similarity-based retrieval for analytical tasks. By incorporating utility signals into the ranking process, the proposed system surfaces passages containing actionable financial evidence.

Adaptive candidate filtering further improves computational efficiency by reducing noise prior to the utility-based ranking stage.








\section{Conclusion \& Future Work}

We present a utility-grounded retrieval architecture for corporate credit underwriting workflows involving long financial documents. The proposed system combines hybrid retrieval, adaptive candidate filtering, and LLM-based utility ranking to prioritize decision-relevant evidence.

Evaluation results demonstrate significant improvements in retrieval accuracy and analyst productivity. The system has been successfully deployed in a large enterprise environment, highlighting the potential of utility-aware RAG architectures for document-intensive decision-support applications.

Future work will explore enhanced structured data extraction from financial tables. The current system preserves traceability when detecting and retrieving tabular content, including document location and structural context. This capability opens the possibility of integrating specialized Optical Character Recognition (OCR) and Vision Language Model (VLM) pipelines to extract structured numerical information directly from table regions. 

By combining table detection that leverages OCR extraction together with VLMs, the system could capture additional financial metrics that are often embedded within complex document layouts. This structured information could then be incorporated into downstream retrieval and reasoning processes, enabling richer analytical queries that combine narrative explanations with precise financial figures.





\section*{Acknowledgments}
This work was conducted at OCBC AI Lab. The authors would like to express their sincere gratitude to Kok Ker Ern Kovan, Lee Sheng Kiat, and Germaine Goh Yanshan for their valuable contributions and support throughout the project. Their insights and domain expertise were instrumental in the successful outcomes presented in this paper. The authors would also like to specifically thank Ren Xuezhe, Lisa Tan, Weiyang Song, Qishuai Zhong, and Kenneth Loh Zhen Xiang for their significant contributions to the successful delivery and shipment of the product. Their dedication and collaboration greatly enhanced the impact and practical application of this work.

\bibliography{biblio}

\end{document}